\ifx\XeTeXversion\undefined
  \pdfoutput=1
\fi
\documentclass[10pt,twocolumn,letterpaper]{article}

\usepackage{cvpr}              

\usepackage{amsmath,amssymb}
\usepackage{booktabs}
\usepackage{multirow}
\usepackage{graphicx}

\usepackage{microtype}
\microtypesetup{stretch=20,shrink=20}

\definecolor{cvprblue}{rgb}{0.21,0.49,0.74}
\usepackage[breaklinks,colorlinks,allcolors=cvprblue]{hyperref}
\hypersetup{
  pdftitle={Lost in Fog: Sensor Perturbations Expose Reasoning Fragility in Driving VLAs},
  pdfauthor={Abhinaw Priyadershi and Jelena Frtunikj},
}

\def\confName{SAIAD}
\def\confYear{2026}

\makeatletter
\newcommand{\copyrightbanner}{%
  \begingroup
  \footnotesize\noindent
  \copyright~2026 IEEE. Personal use of this material is permitted.
  Permission from IEEE must be obtained for all other uses, in any current or
  future media, including reprinting/republishing this material for advertising
  or promotional purposes, creating new collective works, for resale or
  redistribution to servers or lists, or reuse of any copyrighted component of
  this work in other works.\par
  \endgroup
  \vspace{12pt}%
}
\patchcmd{\@maketitle}{\null}{\null\copyrightbanner}{}{}
\makeatother

\title{Lost in Fog: Sensor Perturbations Expose Reasoning Fragility\\
in Driving VLAs}

\author{
Abhinaw Priyadershi\\
NVIDIA Corporation, USA\\
{\small\itshape apriyadershi@nvidia.com}
\and
Jelena Frtunikj\\
NVIDIA GmbH, Germany\\
{\small\itshape jfrtunikj@nvidia.com}
}

\begin{document}

\setcounter{topnumber}{4}
\setcounter{bottomnumber}{2}
\setcounter{totalnumber}{6}
\renewcommand{\topfraction}{0.85}
\renewcommand{\bottomfraction}{0.70}
\renewcommand{\textfraction}{0.10}
\renewcommand{\floatpagefraction}{0.85}
\setlength{\floatsep}{3pt plus 1pt minus 1pt}
\setlength{\textfloatsep}{5pt plus 1pt minus 1pt}
\setlength{\abovecaptionskip}{4pt plus 1pt minus 1pt}
\setlength{\belowcaptionskip}{2pt}

\maketitle
\begin{abstract}
Interpretable autonomous driving planners depend not only on generating
explanations, but also on those explanations remaining reliable under
real-world sensor degradation. In this paper we present a controlled
perturbation study of Vision-Language-Action (VLA) robustness in autonomous
driving, evaluating Alpamayo~R1 (10B parameters) across 1,996 scenarios under
eight sensor perturbations (Gaussian noise at four intensities, two lighting
extremes, and two fog levels; ${\sim}18{,}000$ inference trials). We find
that reasoning consistency is a high-fidelity indicator of trajectory
reliability: when Chain-of-Causation (CoC) explanations change after
perturbation, trajectory deviation spikes $5.3{\times}$ (21.8\,m \vs 4.1\,m),
with $r\!=\!0.99$ across attack types and $r_{pb}\!=\!0.53$ per-sample
(Cohen's $d\!=\!1.12$). A controlled ablation provides evidence that enabling
CoC generation is associated with improved trajectory accuracy (11.8\% on
average across conditions; $p < 0.0001$) under matched inference settings.
Over the tested noise range ($\sigma \in \{10, 30, 50, 70\}$), degradation is
approximately linear ($R^2\!=\!0.957$), while standard input preprocessing
defenses provide only marginal relief. Together, these results establish CoC
consistency as a quantitative proxy for planning safety and motivate
reasoning-based runtime monitoring for safer VLA deployment.
\end{abstract}

\section{Introduction}
\label{sec:intro}

Vision-Language-Action models offer a capability that purely reactive planners
cannot match: they explain their decisions. A VLA like Alpamayo~R1~\cite{alpamayo2024},
the current state of the art for CoC-equipped driving models, does not merely
predict a trajectory but it also states why: ``Slow down because the lead
vehicle is braking ahead in the same lane.'' For engineers certifying autonomous driving
software under ISO~21448 (SOTIF), ISO~PAS~8800 (safety of AI-based systems in road vehicles)~\cite{iso8800_2024}, or NHTSA guidelines, this transparency
matters. A model that articulates its reasoning can be interrogated, audited,
and trusted in ways a black-box planner cannot.

Transparent reasoning offers little practical utility if the underlying model
lacks robustness to sensor degradation. For example, consider a VLA deployed
where morning fog degrades the camera feed. If the model still tracks the lead vehicle
correctly, the explanation should say so. If it loses that cue and silently
pivots to ``Keep lane since the road ahead is clear,'' the trajectory will
follow and the change may go undetected by any external monitor. This is
precisely the failure mode that SOTIF analysis is designed to surface: a
system behaving correctly in nominal conditions but degrading silently when
inputs drift outside the operational design domain (ODD) (e.g.\ driving in
fog is not considered as part of the ODD).

Adversarial robustness research in autonomous driving has mostly focused on
perception components: traffic sign attacks~\cite{eykholt2018robust}, LiDAR
spoofing~\cite{cao2019adversarial}, and camera patches~\cite{tu2020physically}
measure detection accuracy, not trajectory outcomes. VLA robustness
studies~\cite{vla_robust2025,eva_vla2025} target manipulation, not driving.
Recent benchmarks~\cite{robodrivevlm2025} introduced sensor and prompt
corruptions for driving VLMs, and ADvLM~\cite{advlm2024} showed
gradient-based attacks can mislead them, but neither traces the causal chain
from visual corruption through reasoning to trajectory error. That chain is
what we measure.

A systematic analysis of this pipeline yields the four contributions of our
paper:

\begin{enumerate}
    \item \textbf{Chain-of-causation consistency as a trajectory reliability
    signal}: when perturbation changes the model's explanation, deviation
    spikes $5.3{\times}$ (21.8\,m \vs 4.1\,m), a relationship that holds
    both aggregate ($r\!=\!0.99$, $n\!=\!8$ attack types) and per-sample
    ($r_{pb}\!=\!0.53$, point-biserial correlation, $n\!=\!15{,}968$). To our
    knowledge, no prior work has shown that a VLA's explanations can flag
    corrupted predictions.

    \item \textbf{A dose-response curve linking sensor noise to trajectory
    error} ($R^2\!=\!0.957$): ADE rises at 0.0048\,m per unit $\sigma$,
    giving developers a direct formula for deployment risk assessment and
    operational design domain planning.

    \item \textbf{Evidence for benefit of CoC generation}: a controlled
    ablation (same checkpoint, matched decoding settings) shows that enabling
    CoC generation is consistently associated with lower ADE across all tested
    conditions (average 11.8\%; $p < 0.0001$), with larger gains on complex
    maneuvers.

    \item \textbf{Standard input sanitization via preprocessing defenses
    offers limited protection}: six off-the-shelf input-processing methods
    yield only marginal, statistically insignificant relief after Bonferroni
    correction, motivating VLA-specific defense research.
\end{enumerate}

\section{Related work}
\label{sec:related}

\paragraph{Vision-Language-Action Models for Driving.}
VLAs combine large vision-language models with robotic control to produce
explainable driving systems. Earlier end-to-end driving
networks~\cite{e2e_nvidia2016} demonstrated that neural networks can steer
directly from camera pixels, but offered little interpretability.
RT-2~\cite{rt2} showed that web-pretrained VLMs can transfer to robot actions.
Octo~\cite{octo2024} generalized this paradigm to multi-task settings.
DriveLM~\cite{drivelm2024} extended it to driving with graph-structured visual
question answering. Alpamayo~R1~\cite{alpamayo2024} advanced the field with a
10B-parameter architecture generating chain-of-causation explanations alongside
64-waypoint trajectories at 10\,Hz, reporting a 12\% planning improvement in
challenging closed-loop scenarios. Whether that reasoning holds under sensor
degradation remains an open question.

\paragraph{Concurrent VLA Robustness Work.}
Concurrent work has begun to probe VLA robustness in adjacent domains. In
manipulation, Guo~\etal~\cite{vla_robust2025} test 17 perturbation types
across four modalities, identifying the action head as most vulnerable;
Eva-VLA~\cite{eva_vla2025} studies physical variations (table height, lighting
angle) and reports failure rates exceeding 60\%. In driving,
RoboDriveVLM~\cite{robodrivevlm2025} benchmarks VLM trajectory prediction
under six sensor corruptions and five prompt corruptions on
nuScenes~\cite{caesar2020nuscenes}, finding fragility to prompt attacks.
However, these studies largely focus on task-level outcomes; how explanation
stability relates to trajectory error is outside their scope.

\paragraph{Adversarial Robustness in Perception.}
Existing robustness evaluations focus heavily on perception components,
targeting modalities ranging from traffic signs~\cite{eykholt2018robust},
LiDAR~\cite{cao2019adversarial}, cameras~\cite{tu2020physically} to sensor
fusion~\cite{fusion_attack2023}. A critical limitation of these approaches is
their reliance on proxy metrics (e.g., detection AP, segmentation IoU) rather
than system-level planning outcomes. Although recent work has begun to probe
trajectory models either by perturbing input
waypoints~\cite{trajectoryattack} or by applying gradient-based attacks on
VLMs~\cite{advlm2024}, the causal chain from natural sensor degradation,
through reasoning instability, to downstream reasoning failure remains
unexplored. We address this by quantifying the propagation of error from
visual input, through the reasoning bottleneck, to the final trajectory.

\paragraph{Weather Robustness and Safety Evaluation.}
Determining the operational design domain of autonomous systems requires
rigorous stress-testing against environmental corruption. Prior work has
established benchmarks for foggy scene
understanding~\cite{foggy_driving,weatheradapt}, enabled closed-loop weather
simulation via CARLA~\cite{carla_weather}, developed corruption-robustness
protocols~\cite{roboflow_weather}, and contributed datasets such as
KITTI-360~\cite{kitti360} and augmentation strategies~\cite{augmax2021}.
However, these contributions are predominantly evaluated on perception-level
metrics (e.g., semantic segmentation mIoU or detection AP), rather than on
planning outcomes. Thus, they do not quantify how such environmental
corruptions propagate through the complex reasoning chains of
Vision-Language-Action models to impact downstream trajectory fidelity.

\paragraph{Explainability as a Safety Mechanism.}
While the importance of explainable planning is widely
recognized~\cite{explainable_av2021}, traditional approaches rely on post-hoc
mechanisms (e.g., attention visualization~\cite{attention_driving} or saliency
maps~\cite{saliency_av}) that approximate reasoning rather than exposing it.
VLAs fundamentally alter this paradigm by generating intrinsic
Chain-of-Causation explanations directly alongside control outputs. An
unresolved question, however, is whether this reasoning channel remains robust
to perturbations. We address this gap by establishing CoC consistency as a
quantitative proxy for trajectory fidelity (\cref{sec:results}).

\section{Experimental setup}
\label{sec:method}

We evaluate VLA robustness by applying controlled sensor corruptions to real
driving sequences and measuring the resulting degradation in both trajectory
accuracy and reasoning consistency. The evaluation spans 1,996 scenarios
across eight perturbation conditions, and includes a CoC-ablation study,
totaling approximately 18,000 inference trials for the primary evaluation.

\subsection{Model and dataset}
\label{sec:model}

\textbf{Model.} For our experiments we adopt Alpamayo~R1~\cite{alpamayo2024},
a 10B-parameter VLA architecture grounded on the Qwen3-VL vision encoder. The
model processes multi-view camera feeds via dynamic resolution encoding
(164K-197K pixels, aspect-ratio adaptive), fuses visual context with
ego-history to jointly predict a 64-waypoint trajectory at 10\,Hz
($T\!=\!6.4$\,s) and a corresponding chain-of-causation explanation.

\textbf{Dataset.} The evaluation is performed on the
PhysicalAI-Autonomous-Vehicles validation split. We uniformly sample from the
1,996 driving sequences spanning diverse maneuvers, including lane-keeping,
vehicle-following, and unprotected turns. Sample sizes range from $n\!=\!475$
(\textsc{Follow\_Vehicle}) to $n\!=\!14$ (\textsc{Turn\_Left}); to ensure
statistical power, our per-category analysis is restricted to categories with
$n \geq 100$.

\textbf{Baseline.} We compare against a constant-velocity physics baseline
that propagates each clip's recorded ego velocity as a straight-line trajectory
over the 6.4-second horizon, producing 64 uniformly-spaced waypoints with no
steering or acceleration adjustment. This baseline serves to distinguish
learned semantic planning from trivial kinematic continuation; the 68.3\%
improvement it yields should therefore be read as the gap over kinematic
extrapolation, not over learned planners. A more informative comparison
against learned trajectory predictors such as AD-MLP~\cite{zhai2023admlp}
is a natural extension for future work.

\textbf{Scenario taxonomy.} To enable granular robustness analysis, we
stratify the dataset into distinct behavioral categories based on semantic
action primitives extracted from the ground-truth (clean) CoC explanations.
The resulting distribution shows common urban interactions: Vehicle Following
($n\!=\!475$), Intersection Navigation ($n\!=\!344$), Signal Compliance
($n\!=\!302$), and Lane Keeping ($n\!=\!213$). Complex maneuvers such as
Passing ($n\!=\!177$) and Turns (Right/Left, combined $n\!=\!54$) are also
represented, with the remaining tail encapsulated in an Other category
($n\!=\!418$).

\subsection{Perturbation design}
\label{sec:perturbations}

We evaluate robustness across an eight-condition threat model spanning three
corruption modalities (\cref{tab:perturbations}). To preserve geometric
consistency across the VLA's multi-view input, we apply each perturbation
synchronously across all camera views.

\begin{table}[t]
\centering
\caption{Eight perturbation conditions spanning three corruption modalities,
applied synchronously across all camera views. Categories align with the
additive, multiplicative, and physics-based corruption types described in
\cref{sec:perturbations}.}
\label{tab:perturbations}
\resizebox{\columnwidth}{!}{%
\begin{tabular}{@{}llll@{}}
\toprule
\textbf{Category} & \textbf{Name} & \textbf{Implementation} & \textbf{Real-World Correlation} \\
\midrule
\multirow{4}{*}{Sensor Noise}
  & noise\_10 & Gaussian $\sigma\!=\!10$ & Typical CMOS noise \\
  & noise\_30 & Gaussian $\sigma\!=\!30$ & Moderate degradation \\
  & noise\_50 & Gaussian $\sigma\!=\!50$ & Heavy sensor noise \\
  & noise\_70 & Gaussian $\sigma\!=\!70$ & Severe degradation \\
\midrule
\multirow{2}{*}{Photometric Shift}
  & dark   & Brightness $0.4{\times}$ & Tunnel entry, dusk \\
  & bright & Brightness $1.6{\times}$ & Sun glare \\
\midrule
\multirow{2}{*}{Volumetric Scattering}
  & fog\_light & Scattering $\alpha\!=\!0.3$ & Light haze \\
  & fog\_heavy & Scattering $\alpha\!=\!0.7$ & Dense fog \\
\bottomrule
\end{tabular}}
\end{table}

\textbf{Sensor Noise (Additive).} We inject Gaussian noise ranging from
$\sigma\!=\!10$ (simulating typical CMOS read noise~\cite{foi2008practical})
to $\sigma\!=\!70$ (simulating severe low-light gain amplification).

\textbf{Photometric Shift (Multiplicative).} We scale global image intensity
linearly ($0.4{\times}$, $1.6{\times}$) to simulate rapid exposure changes
typical of tunnel entry and direct sun glare, high-risk scenarios
empirically linked to intersection crashes~\cite{nhtsa_glare}.

\textbf{Volumetric Scattering (Physics-based).} We synthesize fog using the
Narasimhan and Nayar atmospheric scattering
model~\cite{narasimhan2002vision}. We calibrate the scattering coefficient
$\alpha$ to meteorological standards: $\alpha\!=\!0.3$ approximates light
haze (${\sim}500$\,m visibility), while $\alpha\!=\!0.7$ approximates dense
fog (${\sim}100$\,m visibility).

\subsection{Evaluation metrics}
\label{sec:metrics}

We quantify system robustness using four complementary metrics that capture
both task performance and reasoning stability:

\begin{enumerate}
    \item \textbf{ADE (Average Displacement Error):} Mean L2 distance between
    predicted and ground-truth waypoints across the 64-step horizon:
    \begin{equation}
        \text{ADE} = \frac{1}{T}\sum_{t=1}^{T}
        \|\hat{\mathbf{p}}_t - \mathbf{p}_t\|_2
    \end{equation}

    \item \textbf{ADE Degradation} ($\Delta$ADE): Increase relative to
    clean-condition prediction.

    \item \textbf{CoC Change Rate:} Fraction of samples whose natural-language
    explanation differs after perturbation. We use exact string matching after
    whitespace normalization; even a single changed word counts.

    \item \textbf{L2 Trajectory Deviation:} L2 norm between clean-condition
    and perturbed predicted trajectories:
    $\|\hat{\mathbf{T}}_{\text{clean}} -
    \hat{\mathbf{T}}_{\text{attack}}\|_2$.
\end{enumerate}

We prioritize ADE over Final Displacement Error (FDE) to capture deviation
dynamics throughout the entire 6.4-second planning window, though FDE trends
remain consistent with ADE across all trials.

\paragraph{Binary \vs Continuous CoC Matching.}
We evaluated three approaches for quantifying explanation change
(binary string matching, sentence-embedding cosine similarity (all-MiniLM-L6-v2),
and Jaccard word overlap) across all 15,968 sample-attack pairs.
Sentence-embedding similarity achieves Spearman $\rho\!=\!{-}0.54$ against
L2 trajectory deviation, comparable to binary point-biserial correlation
($r_{pb}\!=\!0.53$); Jaccard achieves $|r|\!=\!0.45$. We adopt binary matching
for two reasons. This distinction is safety-relevant: for short CoC explanations
(7-30 tokens), a single action-verb change can invert driving intent while
still yielding high embedding similarity (e.g.\ switching from ``keep
distance to the lead vehicle'' to ``keep lane since the road is clear''
yields cosine similarity above 0.85 but reflects fundamentally different
driving intent). Binary matching reliably flags these high-risk semantic
inversions in cases where continuous metrics signal false stability.

\subsection{Reproducibility and compute}
\label{sec:compute}

\paragraph{Threat Model \& Scope.}
This study restricts perturbations to natural sensor degradations: noise, fog,
and lighting extremes, rather than gradient-based adversarial attacks. This
scope reflects three considerations: (1)~environmental corruptions such as
rain, fog, and sun glare occur in routine driving, whereas gradient-based
attacks (FGSM, PGD) remain largely confined to academic settings;
(2)~ISO~21448 (SOTIF) and NHTSA guidelines mandate robustness to operational
design domain conditions, making these perturbations directly relevant to
certification requirements; (3)~deterministic perturbation functions ensure
reproducible benchmarking. Black-box and gradient-based attacks, shown
effective against driving VLMs by Zhang~\etal~\cite{advlm2024}, are deferred
to future work.

\paragraph{Ablation Protocol.}
To disentangle correlation from causation, we conduct a controlled ablation in
which every sample is re-evaluated in a trajectory-only inference mode (CoC
generation suppressed). We use the same model checkpoint, identical decoding
hyperparameters (temperature$\!=\!0.6$, top-$p\!=\!0.98$, seed$\!=\!42$,
fixed via \textsc{torch} CUDA manual seed), and a fixed random seed to evaluate
the effect of explicit reasoning on planning fidelity. Because the seed is
reset identically before clean-condition and perturbed-condition inference,
each comparison is fully deterministic per sample; run-to-run sampling
variance does not contribute to the measured CoC change rate.

\paragraph{Ablation Validity.}
The goal of the CoC ablation is to isolate the effect of generating an
explicit natural-language chain-of-causation from other inference-time
factors. To this end, both ``With CoC'' and ``Without CoC'' runs use the same
model checkpoint, identical visual inputs and perturbations, and identical
decoding hyperparameters, with a fixed random seed to minimize sampling
variance. The only change between conditions is the token generation budget: the
``With CoC'' condition generates up to 512 tokens, allowing full
chain-of-causation output; the ``Without CoC'' condition is limited to
1 token, suppressing all CoC text while leaving model weights, visual
inputs, and decoding hyperparameters unchanged. We
acknowledge that this constitutes a token-budget change rather than a pure
reasoning toggle, and report these findings as evidence under this specific
protocol, while acknowledging that stronger causal identification would
require additional controls (e.g., prompt-level suppression or token-matched
dummy generation) beyond the scope of this paper.

\paragraph{Computational Infrastructure.}
All inference was executed on a single NVIDIA A100-40GB GPU\@. Wall-clock
latency per sample ranges from 8-15\,s, assuming resident model memory and
pre-computed perturbations. The total computational budget is approximately
60 GPU-hours: 40\,hours for the primary evaluation and 20\,hours for the
ablation study. The reduced runtime for the latter case reflects the lower
token generation overhead of the trajectory-only inference mode.

\section{Results}
\label{sec:results}

Under clean conditions, Alpamayo~R1 achieves 2.00\,m ADE ($\sigma\!=\!1.89$)
compared to 6.32\,m ($\sigma\!=\!4.87$) for the constant-velocity baseline,
a 68.3\% improvement (\cref{tab:baseline}; paired $t\!=\!36.97$,
$p < 10^{-257}$). This places the VLA in a fundamentally different
performance regime from naive kinematics, retaining the majority of this
advantage even under severe perturbation.

\begin{table}[t]
\centering
\caption{The VLA reduces ADE by 68.3\% relative to a constant-velocity
baseline on clean inputs ($n\!=\!1{,}996$; $p < 10^{-257}$), establishing
a strong performance floor before any corruption is introduced.}
\label{tab:baseline}
\small
\begin{tabular}{@{}lccc@{}}
\toprule
\textbf{Model} & \textbf{ADE (m)} & \textbf{Std} & \textbf{$p$-value} \\
\midrule
Physics (const.\ velocity) & 6.32 & 4.87 & - \\
Alpamayo R1 (VLA)          & 2.00 & 1.89 & - \\
\textbf{Improvement}       & \textbf{68.3\%} & - & $< 10^{-257}$ \\
\bottomrule
\end{tabular}
\end{table}

\subsection{Perturbation robustness}
\label{sec:perturbation_robustness}

\paragraph{Attack Impact.}
\Cref{tab:overall} ranks all eight perturbations by trajectory
degradation. Heavy noise ($\sigma\!=\!70$) is the clear outlier: $+$0.30\,m
ADE, 52.7\% CoC change rate, and 70.6\% of samples exceeding 5\,m L2
deviation. At the other end, mild noise ($\sigma\!=\!10$) is essentially
invisible to the model ($+$0.01\,m, not significant).

\begin{table}[t]
\centering
\caption{Noise $\sigma\!=\!70$ is the dominant threat
($\Delta\text{ADE}\!=\!+$0.30\,m, 52.7\% CoC flips); mild noise
($\sigma\!=\!10$) is statistically negligible ($p\!=\!0.637$). Eight
conditions ranked by $\Delta$ADE\@, $n\!=\!1{,}996$ each. Cohen's $d$
against the clean baseline ranges $0.005$-$0.149$; the ${>}5$\,m column is the fraction of clips where
$\|\hat{\mathbf{T}}_{\text{clean}} - \hat{\mathbf{T}}_{\text{attack}}\|_2 > 5$\,m,
capturing tail risk more faithfully than mean ADE\@.
$^\dagger p\!>\!0.05$ before correction.}
\label{tab:overall}
\resizebox{\columnwidth}{!}{%
\begin{tabular}{@{}lccccc@{}}
\toprule
\textbf{Attack} & \textbf{ADE} & \textbf{$\Delta$ADE} & \textbf{CoC$\,\Delta$} & \textbf{${>}5$m} & \textbf{$p$} \\
\midrule
Noise $\sigma\!=\!10$           & 2.01 & $+$0.01 & 18.1\% & 18.8\% & 0.637 \\
Dark ($0.4{\times}$)$^\dagger$  & 2.06 & $+$0.05 & 39.0\% & 51.7\% & 0.058 \\
Light Fog ($\alpha\!=\!0.3$)    & 2.06 & $+$0.06 & 16.9\% & 18.1\% & 0.006 \\
Bright ($1.6{\times}$)          & 2.06 & $+$0.06 & 31.9\% & 34.7\% & 0.025 \\
Noise $\sigma\!=\!30$           & 2.07 & $+$0.07 & 35.0\% & 48.5\% & 0.020 \\
Heavy Fog ($\alpha\!=\!0.7$)    & 2.09 & $+$0.09 & 33.8\% & 47.3\% & 0.001 \\
Noise $\sigma\!=\!50$           & 2.16 & $+$0.15 & 45.4\% & 61.0\% & $4{\times}10^{-7}$ \\
Noise $\sigma\!=\!70$           & 2.30 & $+$0.30 & 52.7\% & 70.6\% & $2{\times}10^{-17}$ \\
\bottomrule
\end{tabular}}
\end{table}

The split between CoC change rate and trajectory impact is not uniform:
lighting and fog perturbations cause disproportionately high CoC change
rates relative to their trajectory impact. Dark conditions
flip 39\% of explanations but only degrade ADE by 0.05\,m. This
decoupling (the model re-phrases its reasoning yet still lands near the
right trajectory) is further characterized in the CoC-Trajectory Correlation paragraph below.

\begin{figure}[t]
    \centering
    \includegraphics[width=\linewidth]{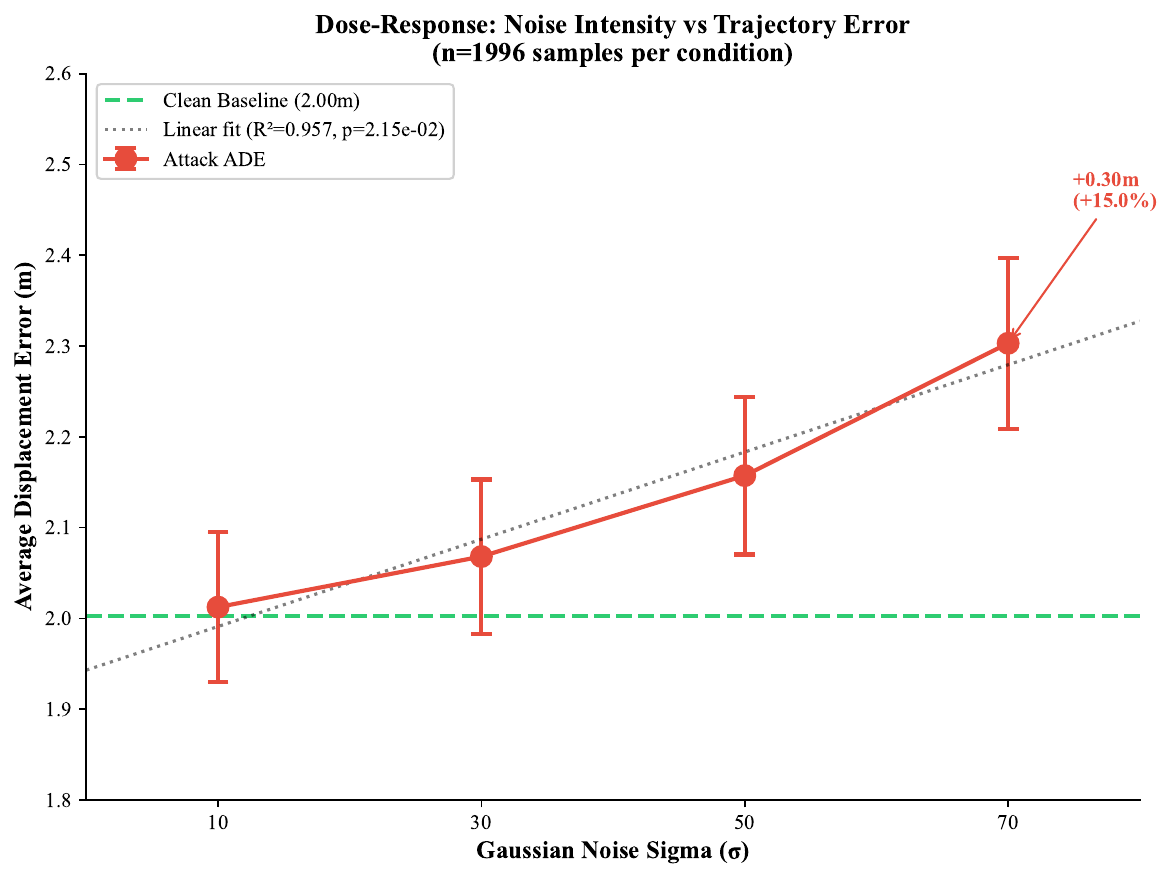}
    \caption{ADE grows linearly with noise intensity ($R^2\!=\!0.957$,
    slope $=\!0.0048$\,m/$\sigma$; four noise levels), enabling a direct
    risk-assessment formula: each 10-unit $\sigma$ increment costs
    $\approx\!0.05$\,m ADE\@. Error bars: 95\% CI\@.}
    \label{fig:dose}
\end{figure}

\paragraph{Dose-Response.}
The four noise levels form a natural dose-response curve
(\cref{fig:dose}). A linear regression on $\sigma$ \vs mean ADE yields
$R^2\!=\!0.957$ ($p\!=\!0.022$), with slope 0.0048\,m per unit $\sigma$.
Each 10-unit $\sigma$ increment adds roughly 0.05\,m to trajectory error.
While moderate ADE increases could partially reflect appropriate conservatism under degraded inputs, the monotonic escalation of the ${>}5$\,m tail from 18.8\% to 70.6\% across noise levels captures large, unpredictable prediction shifts inconsistent with systematic safety behavior.

\begin{table}[t]
\centering
\caption{ADE rises linearly with noise intensity ($R^2\!=\!0.957$), adding
approximately 0.05\,m per 10-unit $\sigma$ increment. Values are mean ADE
with 95\% bootstrap confidence intervals ($n\!=\!1{,}996$ per level).}
\label{tab:dose}
\small
\begin{tabular}{@{}lcccc@{}}
\toprule
$\boldsymbol{\sigma}$ & \textbf{$n$} & \textbf{ADE (m)} & $\boldsymbol{\Delta}$\textbf{ADE} & \textbf{95\% CI} \\
\midrule
10 & 1,996 & 2.01 & $+$0.01 & [1.93, 2.10] \\
30 & 1,996 & 2.07 & $+$0.07 & [1.98, 2.15] \\
50 & 1,996 & 2.16 & $+$0.15 & [2.07, 2.24] \\
70 & 1,996 & 2.30 & $+$0.30 & [2.21, 2.40] \\
\bottomrule
\end{tabular}
\end{table}

Four data points is a thin basis for asserting linearity. Formal model
comparison (AIC) on the four data points favors the linear fit over
log-linear ($\Delta\mathrm{AIC}\!=\!{+}6.3$), power-law
($\Delta\mathrm{AIC}\!=\!{+}6.1$), and saturating
($\Delta\mathrm{AIC}\!=\!{+}8.8$) alternatives, suggesting linearity is
adequate within the tested range, though additional noise levels
($\sigma\!=\!20, 40, 60, 80{+}$) would firm this up considerably. That
said, the practical takeaway stands: degradation is monotonic and the slope
gives a useful rule of thumb for deployment risk assessment.

\paragraph{CoC-Trajectory Correlation.}
\label{sec:coc_correlation}
Partitioning all 15,968 attacked samples ($8 \text{ attacks} \times 1{,}996$
clips) by whether the CoC explanation changed or stayed identical after
perturbation produces the central result of this study.

\begin{figure}[t]
    \centering
    \includegraphics[width=\linewidth]{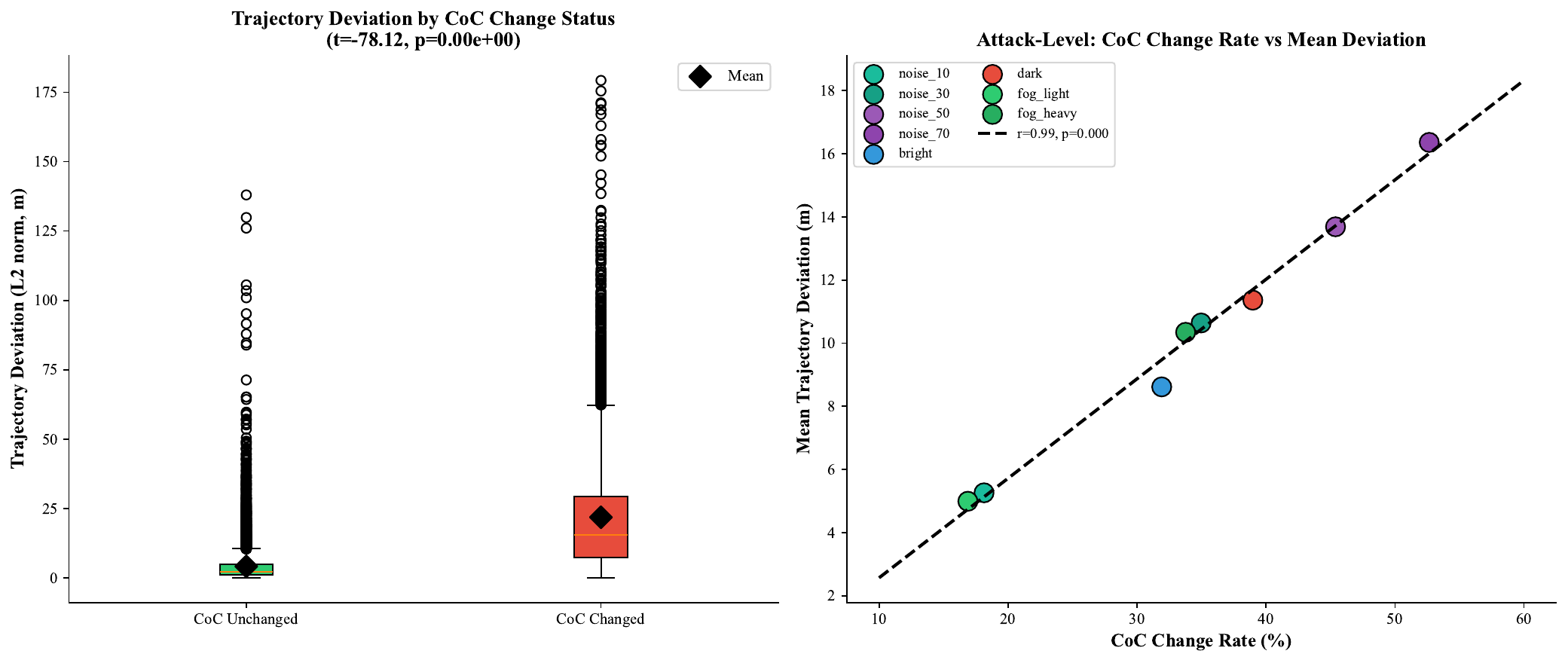}
    \caption{CoC explanation stability is strongly associated with trajectory
    integrity: preserved explanations yield 4.1\,m mean deviation, while
    changed explanations yield 21.8\,m ($5.3{\times}$; $r_{pb}\!=\!0.53$,
    $n\!=\!15{,}968$, left). The relationship holds across attack types
    ($r\!=\!0.99$, $n\!=\!8$, right). Detection characteristics of this
    signal are reported in \cref{tab:monitor}.}
    \label{fig:correlation}
\end{figure}

\begin{table}[t]
\centering
\caption{When a perturbation flips the VLA's CoC explanation, mean trajectory
deviation is $5.3{\times}$ higher than when the explanation is preserved
(21.82\,m \vs 4.13\,m mean; $7.1{\times}$ by median); Cohen's $d\!=\!1.12$,
$p < 10^{-100}$, $n\!=\!15{,}968$.}
\label{tab:coc}
\small
\begin{tabular}{@{}lccc@{}}
\toprule
\textbf{Condition} & \textbf{Mean L2} & \textbf{Median L2} & $\boldsymbol{n}$ \\
\midrule
CoC Unchanged  & 4.13\,m  & 2.16\,m  & 10,525 \\
CoC Changed    & 21.82\,m & 15.39\,m & 5,443  \\
\midrule
Ratio          & $5.3{\times}$ & $7.1{\times}$ & - \\
$t$-test / $p$ & \multicolumn{2}{c}{$t\!=\!{-}78.1$,\quad $p < 10^{-100}$} & - \\
Cohen's $d$    & \multicolumn{2}{c}{$1.12$} & - \\
\bottomrule
\end{tabular}
\end{table}

When the model's explanation stays put, the trajectory barely budges: 4.13\,m
mean (2.16\,m median) deviation. When the explanation flips, deviation jumps
to 21.82\,m mean (15.39\,m median): a $5.3{\times}$ mean ratio and
$7.1{\times}$ median ratio, both robust to the right-skewed deviation
distribution ($t\!=\!{-}78.1$, Cohen's $d\!=\!1.12$). Across attack types,
the correlation is $r\!=\!0.99$ ($n\!=\!8$), indicating that perturbations
that induce more CoC changes produce proportionally larger trajectory errors
with near-perfect linearity (\cref{fig:correlation}). We treat CoC flips as a binary alarm runtime monitoring signal and evaluate its
detection characteristics in our open-loop, paired clean-vs-perturbed setting (\cref{tab:monitor}).

\begin{table}[t]
\centering
\caption{Per-attack point-biserial correlation between CoC stability and L2
trajectory deviation ($n\!=\!1{,}996$ per attack; all $p < 10^{-70}$). Mild
perturbations yield the strongest per-sample diagnostic signal
($r_{pb}\!>\!0.63$, Ratio $>\!12{\times}$). Under severe noise
($\sigma\!=\!70$), even unchanged CoC explanations co-occur with elevated
baseline deviation (8.6\,m), compressing the diagnostic gap to
$2.7{\times}$.}
\label{tab:perpb}
\small
\begin{tabular}{@{}lcc@{}}
\toprule
\textbf{Attack} & $\boldsymbol{r_{pb}}$ & \textbf{Changed/Unchanged} \\
\midrule
Light Fog ($\alpha\!=\!0.3$)  & 0.643 & $13.5{\times}$ \\
Noise $\sigma\!=\!10$         & 0.639 & $12.8{\times}$ \\
Bright ($1.6{\times}$)        & 0.564 &  $7.5{\times}$ \\
Heavy Fog ($\alpha\!=\!0.7$)  & 0.504 &  $4.5{\times}$ \\
Noise $\sigma\!=\!30$         & 0.501 &  $4.6{\times}$ \\
Dark ($0.4{\times}$)          & 0.481 &  $4.1{\times}$ \\
Noise $\sigma\!=\!50$         & 0.438 &  $3.2{\times}$ \\
Noise $\sigma\!=\!70$         & 0.384 &  $2.7{\times}$ \\
\bottomrule
\end{tabular}
\end{table}

\paragraph{CoC as a Runtime Monitor.}
\label{sec:monitor}
Runtime safety monitors are well-studied external mechanisms that operate at
inference time alongside a deployed model, flagging potential safety concerns
without modifying the underlying model weights~\cite{schotschneider2025runtime}.
We evaluate CoC consistency as a binary runtime monitor alarm: an alert is raised
whenever the Chain-of-Causation (CoC) explanation changes after perturbation,
and an event is flagged as unsafe if $\|\hat{\mathbf{T}}_{\text{clean}} -
\hat{\mathbf{T}}_{\text{perturbed}}\|_2 > 5$\,m (matching the tail-risk
threshold in \cref{tab:overall}).

Note: This ``monitor'' is evaluated in an open-loop paired
clean-vs-perturbed protocol; we do not claim an online deployable monitor
in the current setup.
\Cref{tab:monitor} reports precision, recall, false positive rate (FPR\@),
and area under the receiver operating characteristic curve (AUROC\@),
computed from the same per-sample binary outcomes; AUROC uses per-sample
word-similarity as a graded score.
Under mild perturbations (the operationally realistic regime for early
warning), the binary alarm shows strong performance: Noise $\sigma\!=\!10$ achieves
AUROC$\!=\!0.873$ with FPR$\!=\!0.041$, capturing 78.7\% of unsafe events
while raising a false alarm on fewer than 1-in-24 nominal samples. Under
severe noise ($\sigma\!=\!70$), recall falls to 63.3\% and FPR rises to
0.271, reflecting a language-trajectory decoupling in which the trajectory
decoder fails via collapsing kinematic priors while the language branch
continues producing coherent but safety-irrelevant explanations.
We hypothesize this occurs because the language decoder relies on
high-probability text priors that dominate weakened visual tokens,
producing fluent but perceptually ungrounded outputs, while the
trajectory decoder operates more directly on degraded visual features
without such a fallback. Precision
remains high ($\geq\!0.79$) across all conditions: when the monitor does
alarm, it is reliable; the primary safety limitation is recall under severe
degradation.

\begin{table}[t]
\centering
\caption{CoC flip as an alarm signal across all eight conditions (alarm
$=$ CoC changed; unsafe event $=$ L2 deviation $>\!5$\,m,
$n\!=\!1{,}996$ per attack). Precision remains $\geq\!0.79$ throughout;
recall and AUROC degrade under severe noise as language and trajectory
branches decouple.}
\label{tab:monitor}
\small
\begin{tabular}{@{}lcccc@{}}
\toprule
\textbf{Attack} & \textbf{Prec} & \textbf{Recall} & \textbf{FPR} & \textbf{AUROC} \\
\midrule
Noise $\sigma\!=\!10$        & 0.818 & 0.787 & 0.041 & 0.873 \\
Noise $\sigma\!=\!30$        & 0.834 & 0.601 & 0.113 & 0.744 \\
Noise $\sigma\!=\!50$        & 0.838 & 0.624 & 0.189 & 0.717 \\
Noise $\sigma\!=\!70$        & 0.849 & 0.633 & 0.271 & 0.681 \\
\midrule
Dark ($0.4{\times}$)         & 0.842 & 0.635 & 0.128 & 0.754 \\
Bright ($1.6{\times}$)       & 0.791 & 0.728 & 0.102 & 0.813 \\
\midrule
Light Fog ($\alpha\!=\!0.3$) & 0.819 & 0.765 & 0.037 & 0.864 \\
Heavy Fog ($\alpha\!=\!0.7$) & 0.841 & 0.600 & 0.102 & 0.749 \\
\midrule
\textbf{Aggregate}           & \textbf{0.832} & \textbf{0.647} & \textbf{0.102} & \textbf{0.773} \\
\bottomrule
\end{tabular}
\end{table}

\paragraph{Toward Deployable Monitoring.}
The paired clean-vs-perturbed protocol above is an evaluation construct;
in deployment no uncorrupted reference frame is available.
Two strategies can bridge this gap: \textit{temporal consistency}
(comparing CoC explanations across consecutive frames via semantic
similarity, where an abrupt shift absent a corresponding scene change
signals potential degradation) and a \textit{learned surrogate}
(a lightweight classifier on CoC token embeddings trained to predict
elevated trajectory deviation directly, using the paired data in this
study as supervision).
Validating these strategies is an immediate priority for future work.

\subsection{Failure analysis and scenario vulnerability}
\label{sec:failure_scenario}

\paragraph{Qualitative Breakdown.}\looseness=-2
Under severe noise ($\sigma\!=\!70$), changed CoC explanations fall into three
major categories: action-word changes (27.8\%), object-reference changes
(33.4\%), and shifted object focus (22.2\%). Across all eight attack types,
98.9\% of flipped explanations involve at least one action-word or
object-reference change, both categories with direct trajectory implications.
Pure paraphrase flips (changed wording, preserved intent) represent a
negligible fraction, confirming that exact-match sensitivity does not
conflate harmless rephrasing with safety-relevant intent inversions. A representative failure
illustrates the case: ``Keep distance to the lead vehicle because it is
directly ahead'' becomes ``Keep lane to continue driving since no critical
agent needs attention.'' The model fails to track the lead vehicle and
downgrades the scenario from vehicle-following to free cruising, omitting the
necessary deceleration required for safe following.

\begin{sloppypar}\looseness=-1
Critically, perturbed explanations show no calibrated uncertainty: the model
asserts phrases such as ``since the lane ahead is clear'' with essentially
identical confidence under pristine and corrupted inputs. This absence of
epistemic signaling represents a safety-critical gap since reasoning failures
may appear linguistically well-justified even when the underlying perception
has degraded.
\end{sloppypar}

\begin{sloppypar}\looseness=-1
A contrasting pattern under photometric shift is instructive. Dark conditions
($0.4{\times}$ brightness) flip 39\% of CoC explanations, yet ADE degrades by
only 0.05\,m ($p\!=\!0.058$, not significant). This dissociation suggests
CoC instability can register distributional shift before the trajectory decoder
is visibly affected: the reasoning channel pivots while ego-history and
kinematic priors hold the trajectory near its clean-condition path. \Cref{tab:monitor} confirms this: the Dark-condition monitor achieves
AUROC$\!=\!0.754$ with FPR$\!=\!0.128$, detecting 63.5\% of high-deviation
events despite only a 0.05\,m mean trajectory impact, confirming that CoC
instability can flag photometric distribution shift before trajectory metrics
visibly degrade.
\end{sloppypar}

\begin{figure}[t]
    \centering
    \includegraphics[width=\linewidth]{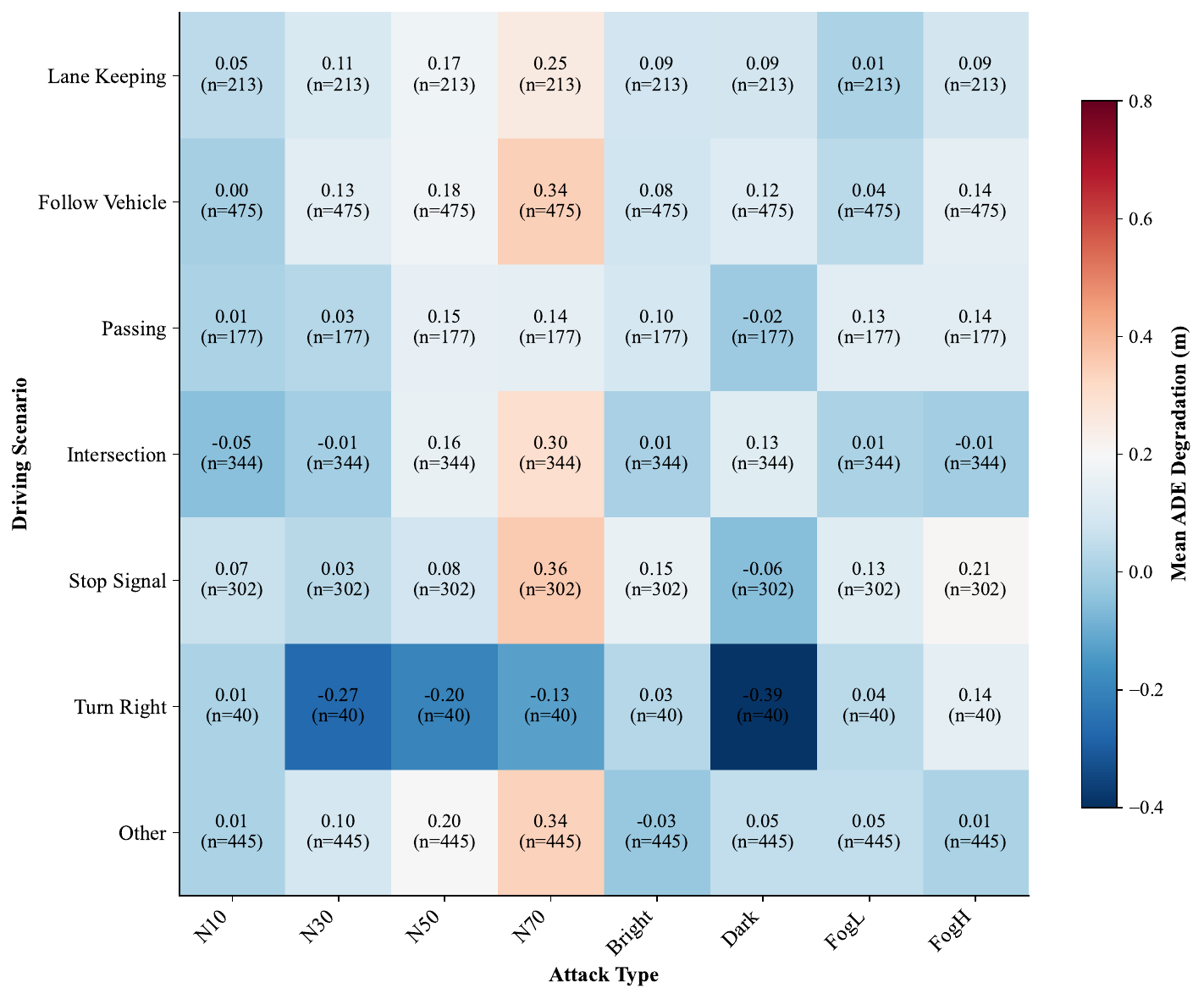}
    \caption{Safety-critical scenarios sustain the greatest degradation across
    all eight attack types: \textsc{Stop\_Signal} ($\Delta\text{ADE}$ up to
    $+$0.36\,m) and \textsc{Follow\_Vehicle} ($+$0.34\,m) are the consistently
    hottest rows regardless of whether the perturbation is noise, lighting, or
    fog. \textsc{Passing} is deceptively resilient in trajectory terms
    ($+$0.14\,m under heavy noise) yet carries the highest CoC change rate
    (58.2\%) among high-$n$ scenarios: another instance of reasoning being
    more fragile than the trajectory it guides.
    $^*$\textsc{Turn\_Right} ($n\!=\!40$) shown for completeness; excluded
    from significance tests ($p\!=\!0.31$).}
    \label{fig:scenario}
\end{figure}

\paragraph{Scenario Vulnerability.}
\Cref{fig:scenario} maps ADE degradation across all eight attack types and
seven scenario categories.

\begin{sloppypar}\looseness=-1
\textsc{Stop\_Signal} and \textsc{Follow\_Vehicle} show the steepest
degradation, which is particularly worrisome given their safety criticality.
\textsc{Passing} scenarios hold up reasonably well in trajectory terms, though
their CoC change rate (58.2\%) is the highest among scenarios with
$n \geq 100$, another example of reasoning being more fragile than the
trajectory itself.
\end{sloppypar}

\begin{sloppypar}\looseness=-2
A counterintuitive pattern reinforces this concern: \textsc{Follow\_Vehicle}
simultaneously holds the lowest clean-condition ADE (1.75\,m, the best
baseline performance) and the highest perturbation sensitivity
($+$0.34\,m). This suggests the model relies most heavily on visual cues in
scenarios where the task is well-defined. Tracking a lead vehicle provides a
strong visual anchor under clean conditions but becomes the specific point of
failure when sensor quality degrades. Scenarios where the VLA performs best
are also the most brittle under distribution shift. Finally, a cross-attack
consistency analysis reveals that 21.4\% of clips fail under three or more
attack types, indicating systematic architectural vulnerability rather than
random sensitivity to individual perturbations.
\end{sloppypar}

\subsection{Ablation and defense baselines}
\label{sec:ablation_defense}

\paragraph{CoC Generation Benefit.}

\begin{table}[t]
\centering
\caption{Enabling CoC generation is associated with reduced ADE by 11.8\% on
average, consistent across every condition including all eight attacks (mean
Cohen's $d_z\!=\!0.14$; all $p < 0.0001$, paired $t$-test with Wilcoxon
confirmation).}
\label{tab:ablation}
\small
\begin{tabular}{@{}lcc|cc@{}}
\toprule
\textbf{Condition} & \textbf{With CoC} & \textbf{W/o CoC} & $\boldsymbol{\Delta}$\textbf{ADE} & \textbf{$d_z$} \\
\midrule
Clean              & 2.01 & 2.31 & $+$0.30 & 0.15 \\
Noise $\sigma\!=\!10$  & 2.02 & 2.25 & $+$0.23 & 0.12 \\
Noise $\sigma\!=\!30$  & 1.97 & 2.34 & $+$0.37 & 0.18 \\
Noise $\sigma\!=\!50$  & 2.10 & 2.42 & $+$0.31 & 0.16 \\
Noise $\sigma\!=\!70$  & 2.23 & 2.53 & $+$0.30 & 0.14 \\
Dark ($0.4{\times}$)  & 2.04 & 2.37 & $+$0.32 & 0.16 \\
Bright ($1.6{\times}$) & 2.03 & 2.26 & $+$0.23 & 0.11 \\
Light Fog          & 2.02 & 2.22 & $+$0.20 & 0.11 \\
Heavy Fog          & 2.09 & 2.34 & $+$0.25 & 0.12 \\
\midrule
\textbf{Average}   & \textbf{2.06} & \textbf{2.33} & $\mathbf{+0.28}$ & \textbf{0.14} \\
\bottomrule
\end{tabular}
\end{table}

\begin{sloppypar}\looseness=-1
Across all nine perturbations, enabling CoC generation reduces ADE
(\cref{tab:ablation}). The average improvement is 11.8\% (range:
9.1-15.8\%); effect sizes are small (mean $d_z\!=\!0.14$) but consistent
without exception. Lane-change maneuvers benefit most ($d_z\!=\!0.43$,
$\Delta\text{ADE}\!=\!0.91$\,m, $p < 0.001$), while simple lane-keeping
benefits least, consistent with the intuition that explicit reasoning is
helpful for complex decisions. CoC-enabled trajectories also degrade 15\%
more slowly under increasing noise (slope 0.0039\,m/$\sigma$ \vs
0.0046\,m/$\sigma$), suggesting reasoning anchors predictions against
distribution shift.
\end{sloppypar}

\paragraph{Preprocessing Defenses.}
We evaluated six off-the-shelf input-preprocessing defenses;
\cref{tab:defense} reports all six alongside the no-defense baseline.

\begin{table}[b]
\centering
\caption{No off-the-shelf preprocessing defense achieves significance after
Bonferroni correction; the best average improvement is 0.7\,m. All six
evaluated defenses are shown, ordered by average L2 improvement. Values are
mean L2 deviation between clean and perturbed predicted trajectories
(not ADE\@); lower $=$ better.}
\label{tab:defense}
\resizebox{\columnwidth}{!}{%
\begin{tabular}{@{}lcccc|c@{}}
\toprule
\textbf{Defense} & \textbf{N\,$\sigma\!=\!30$} & \textbf{N\,$\sigma\!=\!70$} & \textbf{Dark} & \textbf{Fog} & \textbf{Avg\,$\Delta$} \\
\midrule
No Defense            & 21.6 & 23.7 & 21.9 & 21.1 & -      \\
Bilateral             & 20.9 & 22.7 & 20.9 & 21.2 & $+$0.7 \\
Gaussian $3{\times}3$ & 20.8 & 23.0 & 21.4 & 20.9 & $+$0.6 \\
Gaussian $5{\times}5$ & 20.9 & 23.3 & 21.3 & 20.9 & $+$0.5 \\
JPEG Q75              & 21.4 & 23.2 & 21.4 & 20.3 & $+$0.5 \\
Median $3{\times}3$   & 21.0 & 23.4 & 21.4 & 20.8 & $+$0.4 \\
Median $5{\times}5$   & 21.1 & 23.5 & 21.6 & 21.0 & $+$0.3 \\
\bottomrule
\end{tabular}%
}
\end{table}

Improvements range from 0.5-0.7\,m, not statistically significant after
Bonferroni correction.

\paragraph{Severity-Conditioned Effect.}\looseness=-1
A conditional analysis reveals a consequential interaction: preprocessing
defenses reduce L2 deviation by 9.8\,m for samples with severe degradation
(L2 $> 30$\,m) but increase it by 7.9\,m for samples with mild degradation
(L2 $< 10$\,m), likely due to smoothing-induced distribution shift. Since
mild degradation is far more common in practice, a naive ``always apply
defense'' policy is contraindicated and can increase operational risk. These
results motivate adaptive, severity-conditioned defense strategies rather than
uniform preprocessing.

\section{Discussion}
\label{sec:discussion}

\begin{sloppypar}\looseness=-1
While the aggregate ADE degradation appears modest (15\%, from 2.00\,m to
2.30\,m), mean-centric statistics fail to capture the severity of the tail
distribution under sensor degradation: 70.6\% of samples exceed 5\,m L2
deviation under heavy noise, sufficient for multi-lane departure at highway
speeds. In a safety-critical system, this kind of tail risk is often more
consequential than the mean, because rare but large deviations can dominate
operational hazard.
\end{sloppypar}

\begin{sloppypar}\looseness=-2
A related concern is that perturbed explanations provide no signal of input
degradation. The model asserts statements such as ``the lane ahead is clear''
with essentially identical confidence whether the camera feed is pristine or
corrupted, revealing a lack of calibrated epistemic signaling in the language
channel. Therefore, uncertainty-aware language supervision, where models hedge
under degraded inputs, represents a promising training direction.
The language-trajectory decoupling observed under severe noise also
defines the practical boundary of CoC-based monitoring: explanations
serve as a reliable safety proxy in the mild-to-moderate perturbation
regime, where the language branch still tracks perception, but become
less reliable under extreme degradation where text priors dominate.
The coupling between CoC stability and trajectory fidelity (\cref{tab:coc})
is correlational rather than causal: the language and trajectory branches
may share upstream representations without one governing the other, a
trustworthiness gap that certification frameworks such as
ISO~PAS~8800~\cite{iso8800_2024} must address.
\end{sloppypar}

\begin{sloppypar}\looseness=-1
These findings suggest three data augmentation priorities for improving
robustness: heavy Gaussian noise ($\sigma \geq 50$), low-light conditions
(which trigger 39\% CoC change rates despite modest trajectory impact), and
targeted collection for \textsc{Stop\_Signal} and \textsc{Follow\_Vehicle}
scenarios.
\end{sloppypar}

\subsection{Limitations}
\label{sec:limitations}

\paragraph{Methodological Constraints.}\looseness=-1
The perturbations are synthetic and applied independently per frame; real
sensor degradation carries temporal correlations, rolling-shutter artifacts,
and hardware-specific noise characteristics that are not captured in our
stimuli. Moreover, the evaluation is open-loop: each clip receives a corrupted
initial frame with no subsequent feedback, so cascading errors are not captured.

\paragraph{Architectural Generalizability.}\looseness=-1
While Alpamayo~R1 serves as a representative state-of-the-art VLA, the
transferability of these findings to alternative architectures (e.g.,
RT-2, PaLM-E) remains an open empirical question. Similarly, although the
constant-velocity baseline effectively isolates learned reasoning from
kinematic extrapolation, future work should benchmark against learned
trajectory predictors (e.g., Trajectron$++$~\cite{salzmann2020trajectron})
to disentangle VLA-specific vulnerabilities from general forecasting errors.

\paragraph{Statistical \& Computational.}\looseness=-1
The noise dose-response analysis is based on four noise levels; adding
intermediate intensities would strengthen or refute the linearity assumption
($R^2\!=\!0.957$). Similarly, the attack-level correlation ($r\!=\!0.99$) is
computed over only eight attack types; additional perturbation modalities would
confirm whether near-perfect linearity generalizes to a broader threat model.
VLA inference at 10B parameters requires 8-15\,s per frame on an A100 GPU,
far below real-time requirements; the robustness properties of quantized or
distilled variants remain an open research question.

\subsection{Future work}
\label{sec:future}

\begin{sloppypar}\looseness=-2
Building on the monitoring results in \cref{tab:monitor}, promising directions
include temporal CoC consistency checks, perturbation probing at inference time,
and ensemble consensus over CoC outputs, as well as per-scenario breakdown of
operating characteristics beyond the aggregate metrics reported here.
Closed-loop evaluation in a simulation environment (e.g., CARLA, nuPlan)
is a natural next step: it would enable direct collision and off-road rate
computation (not derivable from our open-loop setup, where the ${>}5$\,m
deviation tail serves as a calibrated operational proxy) and capture
cascading errors that accumulate when corrupted predictions feed back into
subsequent frames.
\end{sloppypar}

\section{Conclusion}
\label{sec:conclusion}

\begin{sloppypar}\looseness=-2
In this paper we present a systematic stress-test of Vision-Language-Action
(VLA) robustness in the context of autonomous planning.
By evaluating the Alpamayo~R1 architecture across a stratified set of 1,996
driving scenarios and eight distinct sensor corruptions (totaling
${\sim}18{,}000$ primary inference trials), we quantify the specific vulnerabilities
of end-to-end reasoning models under domain shift. Our analysis shows that
standard input defenses yield no statistically significant improvements, while
performance under Gaussian noise follows a predictable linear degradation
profile ($R^2\!=\!0.957$), providing a calibrated heuristic for deployment
risk. Our primary finding is that reasoning consistency serves as a
high-fidelity proxy for planning safety. We observe a strong coupling between
linguistic stability and trajectory error ($r\!=\!0.99$ across modalities;
$r_{pb}\!=\!0.53$ per-sample), where reasoning collapse signals a
$5.3{\times}$ spike in displacement error (21.8\,m \vs 4.1\,m; Cohen's
$d\!=\!1.12$). A controlled ablation provides evidence consistent with a
benefit of CoC generation: trajectory accuracy is associated with an 11.8\%
improvement across all conditions ($p < 0.0001$).
Together, these results establish CoC consistency as a quantitative proxy for
trajectory fidelity and motivate both VLA-specific defense mechanisms and
reasoning-based runtime monitoring as research directions for improving
autonomous driving safety.
\end{sloppypar}

{
    \small
    \bibliographystyle{ieeenat_fullname}
    \bibliography{main}
}

\end{document}